\documentclass[runningheads]{llncs}

\usepackage{eccv}

\usepackage{eccvabbrv}

\usepackage{graphicx}
\usepackage{booktabs}
\usepackage{makecell}
\usepackage{multirow}
\usepackage{bm}
\usepackage[ruled]{algorithm2e}
\usepackage[accsupp]{axessibility}
\usepackage{amsmath}
\usepackage{amssymb}
\usepackage{booktabs}
\usepackage{tabularx}
\usepackage{multirow}
\usepackage{bbding}
\usepackage{epsfig}
\usepackage{amsmath}
\usepackage{times}
\usepackage{epsfig}
\usepackage{graphicx}
\usepackage{amssymb}
\usepackage{bm}
\usepackage{amsfonts,amssymb}
\usepackage{float}

\usepackage[accsupp]{axessibility}  % Improves PDF readability for those with disabilities.

\usepackage[pagebackref,breaklinks,colorlinks,citecolor=eccvblue]{hyperref}
\usepackage{hyperref}

\usepackage{orcidlink}

\begin{document}

\title{Unrolled Decomposed Unpaired Learning for Controllable Low-Light Video Enhancement}

\titlerunning{Unpaired Low-Light Video Enhancement}

\author{Lingyu Zhu\inst{1} \and
Wenhan Yang\inst{2} \and
Baoliang Chen\inst{1} \and
Hanwei Zhu\inst{1} \and
Zhangkai Ni\inst{3} \and
Qi Mao\inst{4} \and
Shiqi Wang*\inst{1}}

\authorrunning{L.~Zhu et al.}

\institute{City University of Hong Kong, Hong Kong, China \and
PengCheng Laboratory, Shenzhen, China \and
Tongji University, Shanghai, China \and
Communication University of China, Beijing, China \\
\email{Corresponding author: shiqwang@cityu.edu.hk}
}

\maketitle

\begin{abstract}
Obtaining pairs of low/normal-light videos, with motions, is more challenging than still images, which raises technical issues and poses the technical route of unpaired learning as a critical role.
This paper makes endeavors in the direction of learning for low-light video enhancement without using paired ground truth.
Compared to low-light image enhancement, enhancing low-light videos is more difficult due to the intertwined effects of noise, exposure, and contrast in the spatial domain, jointly with the need for temporal coherence.
To address the above challenge, we propose the \textbf{U}nrolled \textbf{D}ecomposed \textbf{U}npaired \textbf{Net}work (\textbf{UDU-Net}) for enhancing low-light videos by unrolling the optimization functions into a deep network to decompose the signal into spatial and temporal-related factors, which are updated iteratively.
Firstly, we formulate low-light video enhancement as a Maximum A Posteriori estimation (MAP) problem with carefully designed spatial and temporal visual regularization.
Then, via unrolling the problem, the optimization of the spatial and temporal constraints can be decomposed into different steps and updated in a stage-wise manner.
From the spatial perspective, the designed Intra subnet leverages unpair prior information from expert photography retouched skills to adjust the statistical distribution. 
Additionally, we introduce a novel mechanism that integrates human perception feedback to guide network optimization, suppressing over/under-exposure conditions.
Meanwhile, to address the issue from the temporal perspective, the designed Inter subnet fully exploits temporal cues in progressive optimization, which helps achieve improved temporal consistency in enhancement results.
Consequently, the proposed method achieves superior performance to state-of-the-art methods in video illumination, noise suppression, and temporal consistency across outdoor and indoor scenes.
Our code is available at \url{https://github.com/lingyzhu0101/UDU.git}

\end{abstract}

\section{Introduction}
\label{sec:intro}
Recently, there has been a significant interest in low-light image and video enhancement, which focuses on enlightening images and videos captured in low-light conditions. 
The objective is to make the hidden information in the dark regions more visible, offering visually pleasing results with well-lit illumination, balanced color, and suppressed noise. 
This area of research has gained attention in various emerging computer vision domains, including object detection \cite{yang2020advancing}, autonomous driving\cite{li2021deep}, and facial recognition\cite{kamenetsky2018image}.

Although deep learning-based image enhancement methods have shown promising results in reference-free strategies~\cite{guo2020zero, jiang2021enlightengan, liu2021retinex}, those efforts for videos have received limited attention.
When directly applying these above-mentioned image-based methods to handle videos, two issues arise.
The first major issue is the neglect of temporal contexts among adjacent frames in video enhancement when processing low-light images independently on each frame, which can lead to inconsistencies in the enhanced video frames.
The other main limitation of those methods is that, as they only learn from the aligning distributions, therefore might lead to overexposure or underexposure as the absence of pixel-wise constraint and human perception feedback.
As such, a desirable robust low-light video enhancement should address the following issues:  (i) it can learn restoration knowledge from the unpaired dataset; (ii) besides obtaining good spatial statistical properties, it can also be good at maintaining temporal consistency; and (iii) it owns the mechanism to prevent over/under-exposure.
The reason for meeting these problems lies in that, the temporal and spatial domain degradation are intricately interwoven.
When pixel-wise supervision is absent, the model fails to disentangle the degradation related to spatial and temporal domains, where specially designed constraints can provide useful guidance for suppressing that visual degradation.

Recently, the unrolling methods, which construct deep networks by formulating the related visual prior into a MAP problem and unrolling it into a progressive solution, offer both superior performance with excellent interpretability.
Different from purely deep learning-based methods, which heavily rely on large-scale datasets and powerful computational resources to learn capable latent representations,
the unrolling method provides a more explainable solution and decomposes the optimization into several sub-problems, which can be solved independently and easily.
For example, deep plug-and-play (PNP) methods incorporate pre-trained CNN denoisers as priors within iterative optimization frameworks for inverse imaging problems~\cite{meinhardt2017learning} and snapshot compressive imaging~\cite{yuan2020plug}.
For low-light image enhancement, there are also recent advancements based on deep unfolding networks~\cite{ma2022toward, liu2021retinex, wu2022uretinex}.
These results inspire us to explore the deep unfolding path and build deep networks based on the unrolling form of an optimization function.
In this way, the complex degradation in low-light conditions simplifies the optimization function and allows us to tackle the video enhancement problem from both the spatial and temporal perspectives.
However, existing methods under this route still have room to be improved.
In~\cite{wu2022uretinex}, the separate training of individual modules instead of an end-to-end training approach tends to result in converging to a local optimum.
The limitation of neglecting the true underlying reflectance based on Retinex-inspired optimization unrolling, as stated in~\cite{liu2021retinex}, results in an inaccurate mutual connection between reflectance and illumination.
Similarly, in \cite{ma2022toward}, the approach fails to remove noise, making it only suitable for the case of low sensitivity to noise or when the degradation is known.

\begin{figure}[t]
\centering
\begin{subfigure}{0.9\textwidth}
  \centering
  \captionsetup{justification=centering}
  \includegraphics[width=\linewidth]{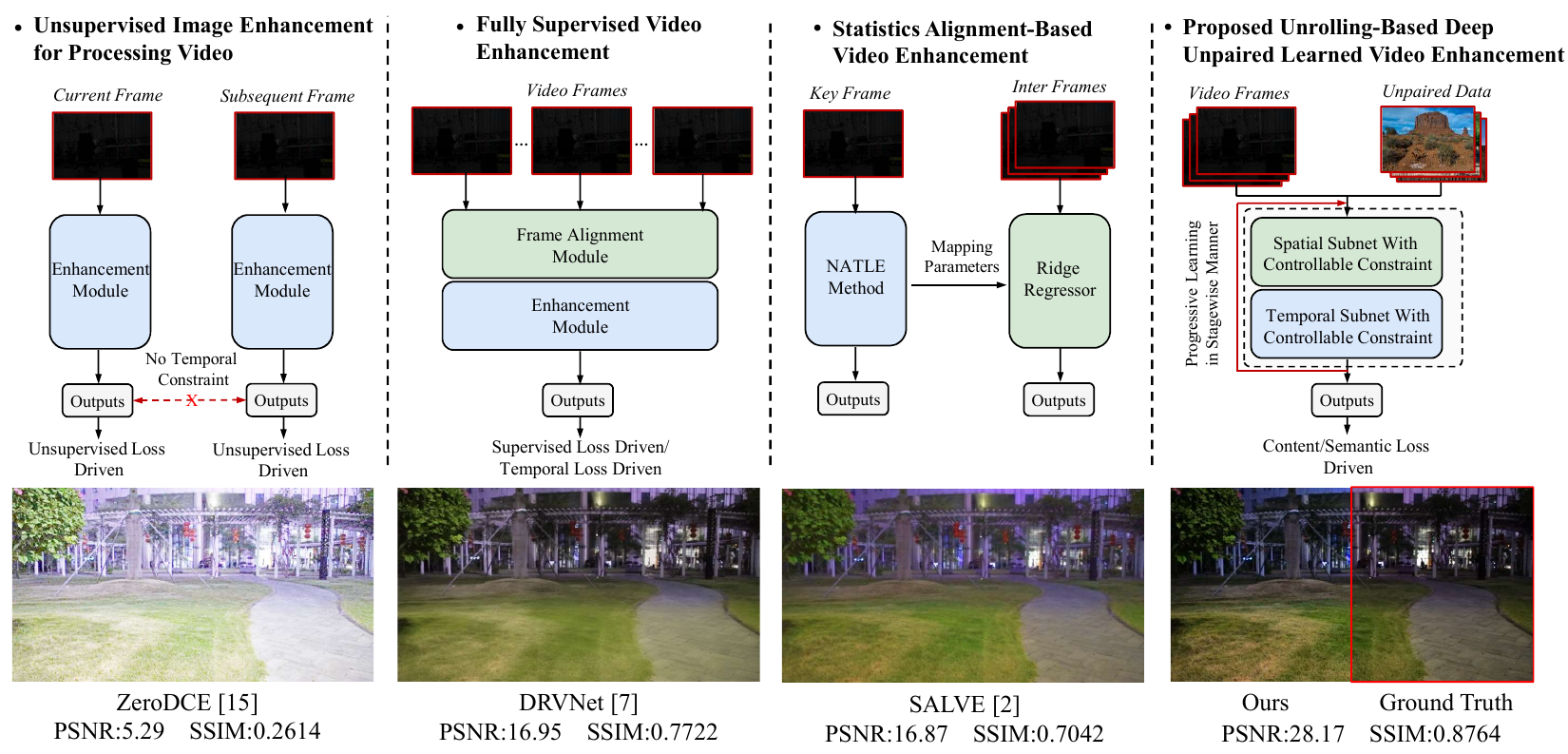}
  \label{fig:img2}
\end{subfigure}
\caption{Visual comparison of the proposed method, existing methods, and the ground truth (highlighted in red box).
}
\label{fig: qualitative compare with existing method}
\end{figure}

To the best of our knowledge, the integration of progressive optimization has not been investigated in the context of deep unfolding architecture for low-light video enhancement.
In this case, the degradation processes under low-light conditions are complex and uncertain, characterized by time-varying, signal-dependent, and spatially variant distributions.
To address these issues and offer interpretability to achieve the controllable enhancement, we propose a novel Unrolled Decomposed Unpaired Network (UDU-Net) without reliance on the reference guidance.
In detail, the spatial and temporal visual priors are jointly injected into a MAP optimization function, 
which further decomposes the restoration process towards these two directions progressively to suppress their degradation and their intertwined one.
This optimization is further unrolled into an end-to-end trainable deep network, which is capable of handling complex real-world degradation.
Our method outperforms existing unpaired low-light enhancement methods.
Notes that, the proposed method even achieves comparable/superior performance to the captured reference in certain cases, as shown in Fig.~\ref{fig: qualitative compare with existing method}.
To summarize, this work has the following contributions:
\begin{itemize}
\item  A trainable deep unfolding video enhancement model is proposed. 
This is the first attempt to build an interpretable low-light video enhancement model that is unrolled to a progressive process from a MAP optimization function but is capable of learning complicated mapping in a data-driven manner, with the constraint from both spatial and temporal perspectives jointly.
The proposed method is designed based on the reference-free strategy and gets rid of the need for paired data.

\item We introduce the controllable human-perception feedback to suppress over/under-exposure under the supervision of quality feedback.
The designed Intra subnet achieves learning the multi-granularity illumination from an initial coarse illumination representation estimated from unpaired learning and incorporates finer illumination progressively under human perception feedback.

\item We explore making full use of temporal cues derived from the intrinsic temporal correlation, which facilitates learning to estimate textural details from both spatial and temporal clues.
In addition, the guided side priors, \textit{i.e.} occlusion masks and optical flows, are injected into the optimization for improving temporal consistency.

\item Extensive experiments demonstrate that the proposed method can fulfill the low-light video enhancement task and provide enhancement results with superior visual quality in both realistic indoor and outdoor scenarios.

\end{itemize}

\section{Related Work}
\subsection{Low-light Image Enhancement}
\noindent \textbf{Traditional methods.}
There are generally two main categories of traditional image enhancement methods: histogram equalization (HE) methods \cite{arici2009histogram, lee2013contrast} and Retinex-based methods \cite{wang2013naturalness, fu2016weighted}.
HE methods strengthen the histogram of given low-light images to increase dynamic and contrast. 
However, one drawback of HE methods is that they can inadvertently amplify hidden noise present in the image.
Retinex-based methods assume that the input image is composed of reflectance and illumination layers. 
By independently manipulating the above two components, these methods aim to improve both the contrast and the overall appearance of the image. 

\noindent \textbf{Deep Learning-based methods.}
The boom in powerful deep-learning representations has propelled learning-based methods to the forefront, offering promising solutions for low-light conditions.
The learnable context map was investigated in low-light image enhancement~\cite{zhu2022enlightening}.
Yang \textit{et al.} \cite{yang2020fidelity} proposed a novel approach called the deep recursive band network (DRBN) that combines the strengths of fully supervised and unsupervised learning methodologies, which leverage fidelity and perceptual quality to improve the appearance of low-light images.
Furthermore, unsupervised learning techniques have been investigated in the context of enhancing low-light images.
To enhance the generalization capability, EnligthenGAN was proposed by Jiang \textit{et al.} \cite{jiang2021enlightengan}, employing a dual-discriminator to achieve global and local improvement.
The Zero-DCE method, introduced in the work of \cite{guo2020zero}, addressed the task of light enhancement by formulating it as an image-specific curve estimation problem.
Liu \textit{et al.} \cite{liu2021retinex} introduced a framework that combines the principled optimization unrolling technique with a cooperative reference-free learning strategy.

\subsection{Low-light Video Enhancement}
\noindent \textbf{Traditional methods.}
Traditional video enhancement methods can be broadly categorized into two groups: tone-mapping-based methods~\cite{bennett2005video,malm2007adaptive,kim2015novel} and physical model-based methods \cite{wang2014piecewise,liu2016low,dong2010fast}.
Tone-mapping-based methods are employed to increase the brightness of low-light videos using tone-mapping algorithms. 
However, these methods tend to incur a relatively high computational cost due to the need for multiple enhancement operations performed in several steps.
The physical model-based methods in low-light video enhancement are often designed based on the Retinex model \cite{land1977retinex} and the atmospheric scattering model \cite{mccartney1976optics}, which depend on accurately estimating illumination and transmission maps, respectively.

\noindent \textbf{Deep Learning-based methods.}
The rapid advancement of deep learning has facilitated the development of learning-based techniques for enhancing low-light videos. 
Zhu \textit{et al.} \cite{zhu2024temporally} designed one video temporal consistency framework to enhance low-light videos.
%
%Lv \textit{et al.} \cite{lv2018mbllen} proposed the approach called the Multi-Branch Low-Light Enhancement Network (MBLLVEN), which introduced a decomposition strategy to address the challenges of low-light video enhancement by breaking down the problem into sub-problems at different levels.
%
StableLLVE \cite{zhang2021learning} is a method that focuses on enforcing temporal stability in low-light video enhancement using only static images. 
It achieves this by leveraging optical flow techniques to mimic the motions typically observed in dynamic scenes. 
In \cite{fu2023dancing}, the Light Adjustable Network (LAN) was proposed to leverage the Retinex-based approach to enhance low-light videos. 
Wang \textit{et al.}~\cite{wang2021seeing} developed a mechatronic system to capture high-quality video pairs under both low-light and normal-light conditions.

\section{Methodology}
\label{sec:method}
\subsection{Low-light Video Modeling}
In low-light scenarios, the captured video is affected by different types of noise, including photon shot noise,  banding pattern noise, read noise, and quantization noise~\cite{wei2020physics}, \textit{etc}.
We can express the imaging process as follows,
\begin{footnotesize}
\begin{equation} \label{low light modeling}
\boldsymbol{y}_{i} = \mathbf{A}_{i} \boldsymbol{x}_{i} + \boldsymbol{n}_{i}, i= 1,2, \ldots, t \ldots, N,
\end{equation}
\end{footnotesize}
wherein, $N$ denotes the total number of frames, $\boldsymbol{y}$ represents the observed video, $\boldsymbol{x}$ is the true underlying video, $\mathbf{A}$ is the degraded matrix, $\boldsymbol{n}$ represents the signal-dependent mixture noise.
Technically, solving Eq.~\eqref{low light modeling} inversely can be formulated as a Bayesian estimation problem~\cite{buades2005non, dabov2007image, mou2022deep}, which can be solved within a unified Maximum A Posteriori (MAP) framework,
\begin{footnotesize}
\begin{align} \label{data fidelity and prior term}
\hat{\boldsymbol{x}} & = \underset{\boldsymbol{x}}{\operatorname{argmin}} \frac{1}{2}\|\boldsymbol{y}-\mathbf{A} \boldsymbol{x}\|_2^2+ \lambda_s J_s(\boldsymbol{x}) + \lambda_t J_t(\boldsymbol{x}),
\end{align}
\end{footnotesize}
where $J( \cdot )$ is the regularizer representing the visual prior preference, $ J_s( \cdot )$ is the spatial regularizer, $ J_t( \cdot )$ is the temporal regularizer, and $\lambda$ is a weighting parameter.
By introducing an auxiliary variable  $\boldsymbol{u}$ and $\boldsymbol{v}$, we obtain,
\begin{footnotesize}
\begin{equation} \label{data fidelity and prior term}
(\hat{\boldsymbol{x}}, \hat{\boldsymbol{u}}, \hat{\boldsymbol{v}}) = \underset{\boldsymbol{x}, \boldsymbol{u}, \boldsymbol{v}}{\operatorname{argmin}} \frac{1}{2}\|\boldsymbol{y}-\mathbf{A} \boldsymbol{x}\|_2^2+ \lambda_s J_s(\boldsymbol{u}) + \lambda_t J_t(\boldsymbol{v}), \text { s.t. } \boldsymbol{x}=\boldsymbol{u} \text{, } \boldsymbol{x}=\boldsymbol{v}.
\end{equation}
\end{footnotesize}

In~\cite{chan2016plug}, the ADMM technique converts the constrained problem into subproblems as follows,
\begin{footnotesize}
\begin{align} 
\label{two subproblems}
\boldsymbol{x}^{k+1}&=\underset{\boldsymbol{x}}{\operatorname{argmin}} \frac{1}{2}\|\boldsymbol{y}-\mathbf{A} \boldsymbol{x}\|_2^2 +\frac{\mu}{2}\left\|\boldsymbol{x}-\widetilde{\boldsymbol{x}}^{k}_s \right\|_2^2 +\frac{\mu}{2}\left\|\boldsymbol{x}-\widetilde{\boldsymbol{x}}^{k}_t \right\|_2^2
\\
\boldsymbol{u}^{k+1}&=\underset{\boldsymbol{u}}{\operatorname{argmin}} \lambda_s J_s(\boldsymbol{u}) +\frac{\mu}{2}\left\| \boldsymbol{u}-\widetilde{\boldsymbol{u}}^{k}_s \right\|_2^2, \overline{\boldsymbol{y}}^{k+1} = \overline{\boldsymbol{y}}^{k} + \left( \boldsymbol{x}^{k+1} - \boldsymbol{u}^{k+1}\right),
\\
\boldsymbol{v}^{k+1}&=\underset{\boldsymbol{v}}{\operatorname{argmin}} \lambda_t J_t(\boldsymbol{v}) +\frac{\mu}{2}\left\| \boldsymbol{v}-\widetilde{\boldsymbol{v}}^{k}_t \right\|_2^2, \overline{\boldsymbol{z}}^{k+1} = \overline{\boldsymbol{z}}^{k} + \left(\boldsymbol{x}^{k+1} - \boldsymbol{v}^{k+1}\right),
\end{align}
\end{footnotesize}
where $\mu$ is a parameter related to $\lambda$.
$ \widetilde{\boldsymbol{x}}^{k}_s = \boldsymbol{u}^{k} - \overline{\boldsymbol{y}}^{k} $,
$ \widetilde{\boldsymbol{x}}^{k}_t = \boldsymbol{v}^{k} - \overline{\boldsymbol{z}}^{k} $,
$ \boldsymbol{u}^{k}             = \overline{\boldsymbol{y}}^{k} + \boldsymbol{x}^{k+1}$, 
$ \boldsymbol{v}^{k}               = \overline{\boldsymbol{z}}^{k} + \boldsymbol{x}^{k+1}$,
$ \overline{\boldsymbol{y}}^{k}    = (1/\rho) \boldsymbol{y}^{k} $,
$ \overline{\boldsymbol{z}}^{k}    = (1/\rho) \boldsymbol{z}^{k} $.
The $\boldsymbol{x}$-subproblem is a simple quadratic optimization that admits a closed-form solution, more details can refer to~\cite{dong2018denoising, mou2022deep}.
The solvers of the $\boldsymbol{u}$ and $\boldsymbol{v}$-subproblems can be low-light enhancement or noise suppression methods that focus on a single image and multiple frames, respectively.
The $\boldsymbol{y}$, $\boldsymbol{z}$-subproblems and the related variables are solved from a linear solver, which can be easily simulated by neural network components.
This derivation and analysis demonstrate a critical insight:
\textit{the solution of this MAP problem can be unrolled into the cascaded single-image enhancer and multi-frame enhancer, connected by simple neural network components for video low-light enhancement}.
\textit{Their focus on different stages can eventually forge into comprehensive considerations on the degradation from the spatial, and temporal aspects and their interweaved ones.}

\begin{figure*}[t]
\centering
\includegraphics[width=0.80\linewidth]{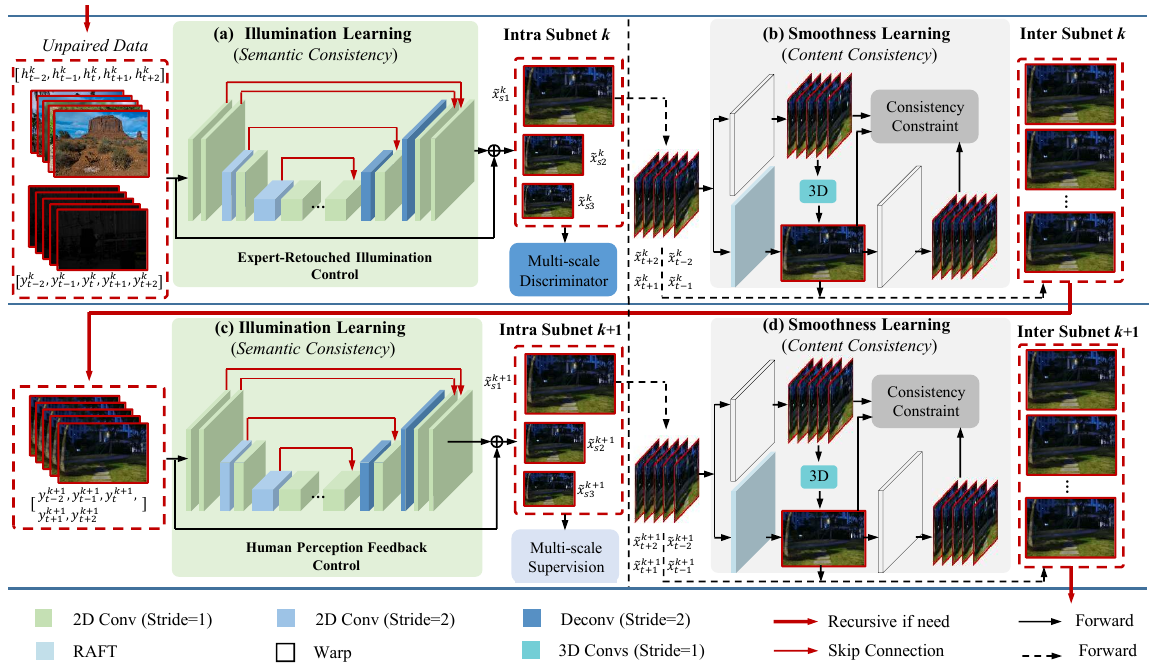}
\caption{
Illustration of the proposed Unrolled Decomposed Unpaired Network (UDU-Net), which is derived from the concept of unrolling and employs a stage-by-stage manner from spatial and temporal perspectives to effectively separate intertwined degradation.
Both spatial and temporal visual priors are incorporated into the process, where the spatial prior incorporates unpaired retouched illumination and human perception feedback, while the temporal prior involves exploring temporal cues and utilizing guided side information.
}
\label{main framework}
\end{figure*}

\subsection{Discussion on Spatial and Temporal Regularizer $J_s( \cdot )$ and $J_t( \cdot )$ }
Our work pays attention to restoring videos from both spatial and temporal degradation in low-light environments and 
constructs sub-components that suppress this degradation with spatial and temporal regularizers.

\noindent \textbf{a) Spatial Visual Prior}: \noindent \textbf{Unpaired Retouched Illumination.}
Expert-retouched illumination is closely associated with desired illumination characteristics, as professional photographers often make precise adjustments based on global information such as exposure, vivid colors, high contrast, and local textures, \textit{etc}.
\textit{By estimating the current low-light frame using an unpaired expert-retouched dataset, we can accurately estimate the illumination distribution in a manner that aligns with the preferences of general users.}
Herein, we adopt MIT-Adobe FiveK Dataset~\cite{bychkovsky2011learning} retouched by expert C to leverage the expertise of the retouching process and significantly improve the overall visual appearance of the enhanced frames.
\noindent \textbf{Human Perception Feedback.} 
However, once the model has learned the illumination distribution from the high-quality version dataset, further improvement becomes challenging due to the limitations imposed by the dataset serving as an upper bound.
To ensure a stagewise optimization process, it is essential to consider multi-granularity illumination beyond intrinsic high-quality version dataset stage by stage, allowing for progressive learning and improving the overall enhancement results.
\textit{Hence, it is beneficial to introduce controllable intrinsic illumination, which is better than the unpaired expert-retouched dataset.}
To prevent failure cases like over-exposure, we incorporate illumination regularization using controllable human perception feedback, which is guaranteed by the proxy human visual system evaluation, ensuring visual enhancements in line with human perception.

\noindent \textbf{b) Temporal Visual Prior:} \noindent \textbf{Temporal Cues Exploration.}
The true underlying video should exhibit high consistency and correlation along the temporal dimension, due to the continuous nature of background contexts.
As such, well-aligned frames and minimal differences can be achieved by exploring the enhanced temporal information.
However, even if good motion estimation and compensation are achieved, the presence of obstinate noise can still affect the alignment of the background and object contexts due to the existence of annoying artifacts, such as complex noise and uneven exposure, \textit{etc}. 
\textit{Hence, it is beneficial to remove irrelevant factors by forcing the model to generate temporal coherence results.}
Our work incorporates motion estimation into the optimization target to reduce annoying noise and reduce brightness difference, alleviate temporal artifacts, and enhance the coherence of the output video.
\noindent \textbf{Guided Side Information.}
Besides the aforementioned constraints, we introduce another valuable side information to guide the video enhancement process. 
This includes incorporating independent signals like mask maps as part of the loss functions, enabling a tradeoff between texture preservation and selective noise removal throughout the context region.

\section{Controllable Low-light Video Enhancer}
\subsection{Network Architecture}
Inspired by Sec.~\ref{sec:method}, we propose the novel UDU-Net as shown in Fig.~\ref{main framework}.
It takes a progressive architecture consisting of an Intra subnet and an Inter subnet connected by neural network components.
In the stage-wise process, the model optimizes toward improving spatial and temporal visual properties progressively, which leads to improved visual performance.

\noindent \textbf{Intra Subnet}.
This part focuses on illumination learning by extracting expert-retouched illumination prior from unpaired data (stage $\textit{k}$) and leveraging human perception priors under intrinsic illumination constraints (stage $\textit{k+1}$). 

\begin{itemize}
\item  
In stage $\textit{k}$, subnetwork (a) (Fig.~\ref{main framework} (a)) of the framework focuses on estimating normal-light characteristics from the distribution of unpaired high-quality images, enabling adjustments in contrast and brightness while maintaining content and intrinsic semantic consistency between input and enhanced frames.

\item In stage $\textit{k+1}$, subnetwork (c) (Fig.~\ref{main framework} (c)) focuses on achieving a refined illumination level, guided by human perception feedback, in comparison to the initial coarse illumination obtained from the previous stage.
This leads to improved enhancement results, as the achievement of a better illumination level is considered an integral part of our overall optimization target. 
\end{itemize}

\noindent \textbf{Inter Subnet}.
Both Fig.~\ref{main framework} (b) and Fig.~\ref{main framework} (d) undergo the same process, which focuses on learning temporal consistency through fully utilizing intrinsic temporal cues and effectively suppressing noise by applying a manually set mask threshold for filtering.

\begin{itemize}
\item The utilization of 3D convolution layers in the proposed method allows for the prediction of the noise-free layer of the current frame by merging information from its adjacent video frames. 
This enables the model to maintain temporal consistency and improve the overall quality of the current frame.

\item Guided by the noise-free estimation through 3D convolution, we further enhance the details using a network taking the same architecture.
This network takes the current frame and adjacent video frames as input and generates the residual details while maintaining inter-frame consistency.
To prevent inter-frame consistency from the effect of noise, the mask map is incorporated into the loss function.

\end{itemize}

\subsection{Controllable Propagation Mechanism}
\subsubsection{Illumination Optimization.}
We first introduce a mapping $G_{\boldsymbol{\theta}} \left(\cdot\right)$  with parameters $\theta$ to learn the illumination. The overall unit is written as,
\begin{footnotesize}
\begin{equation} \label{Intra Net Optimization}
{F}_{\text {Intra}}\left(\boldsymbol{x}^k\right):\left\{\begin{array}{l}
\boldsymbol{r}_{\text{intra}}^k=G_{\boldsymbol{\theta}}^k\left(\boldsymbol{x}^k\right), \boldsymbol{x}^0=\boldsymbol{y} \\
\widetilde{\boldsymbol{x}}^{k}=\boldsymbol{x}^k + \boldsymbol{r}_{\text{intra}}^k,
\end{array}\right.
\end{equation}
\end{footnotesize}
wherein $\boldsymbol{r}_{\text{intra}}^k$ and $\widetilde{\boldsymbol{x}}^k$ denote the residual term and enhanced output at the stage $\textit{k}$. 

\noindent \textbf{Stage $\textit{k}$}:
Technically, the illumination can be obtained by minimizing the following regularized energy function,
\begin{footnotesize}
\begin{equation} \label{illumination fidelity k}
\widetilde{\boldsymbol{x}}^{k}=\underset{\boldsymbol{x}}{\operatorname{argmin}} \frac{1}{2}\left\|\boldsymbol{x}-\left(\widetilde{\boldsymbol{x}}^{k}-\rho \nabla F_{\text {Intra}}\left(\widetilde{\boldsymbol{x}}^{k}\right)\right)\right\|_2^2.
\end{equation}
\end{footnotesize}

The algorithm aims to find the optimal solution that minimizes the objective function defined by the illumination term $F(\cdot)$. 
$\nabla$ is the differential operator.
The step size $\rho$ determines the size of each update.

\noindent \textbf{Loss Function}: The semantic self-supervised loss plays a crucial role in maintaining semantic information consistency, which is defined as the  $\boldsymbol{l}_{2}$  norm between the feature maps of the input low-light video frames and those of the generated high-quality video frames as follows,
\begin{footnotesize}
\begin{equation} \label{content self-supervised loss}
\begin{aligned}
& \mathcal{L}_{\text {semantic-G}}=\sum_{i=t-2}^{t+2}\sum_{j=1}^{J}\left\|   \phi_j\left( G_{\boldsymbol{\theta}}^k\left(\boldsymbol{y}_i^k\right) + \boldsymbol{y}_i^k \right)   -  \phi_j\left( \boldsymbol{x}_{i}^{k} \right)    \right\|_2^2.
\end{aligned}
\end{equation}
\end{footnotesize}

Herein, $\phi_j$ denotes the extracted $j$ feature layer using the pre-trained VGG network~\cite{simonyan2014very}.
In addition, the content self-supervised loss to maintain the unpair high-quality images ($\boldsymbol{h}_{i}^{k}$) unchanged in the three scales (S=3), which is defined as follows,
\begin{footnotesize}
\begin{equation} \label{content self-supervised loss}
\begin{aligned}
& \mathcal{L}_{\text {content-G}}=\sum_{i={t-2}}^{t+2}\sum_{s=1}^{S}\left\| \left(  G_{\boldsymbol{\theta}}^k\left(\boldsymbol{h}_i^k\right) + \boldsymbol{h}_i^k\right) -  \boldsymbol{h}_{i}^{k}  \right\|_1.
\end{aligned}
\end{equation}
\end{footnotesize}
The Relativistic Average HingeGAN \cite{jolicoeur2018relativistic} is adopted to train discriminator $D_{\boldsymbol{\theta}}(\cdot)$ to assess whether the generated illumination distribution has been effectively learned.

\noindent \textbf{Stage $\textit{k+1}$}: Similarly, the stage $\textit{k+1}$ needs to be defined as an optimization target and formulated by the universal descent-direction-based scheme.
\begin{footnotesize}
\begin{equation} \label{illumination fidelity k+1}
\widetilde{\boldsymbol{x}}^{k+1}=\underset{\boldsymbol{x}}{\operatorname{argmin}} \frac{1}{2}\left\|\boldsymbol{x}-\left(\widetilde{\boldsymbol{x}}^{k+1}-\rho \nabla F_{\text {Intra}}\left(\widetilde{\boldsymbol{x}}^{k+1}\right)\right)\right\|_2^2.
\end{equation}
\end{footnotesize}

\begin{figure*}[t]
\centering    
\subfloat[Score:27.52] {\includegraphics[width=0.20\linewidth]{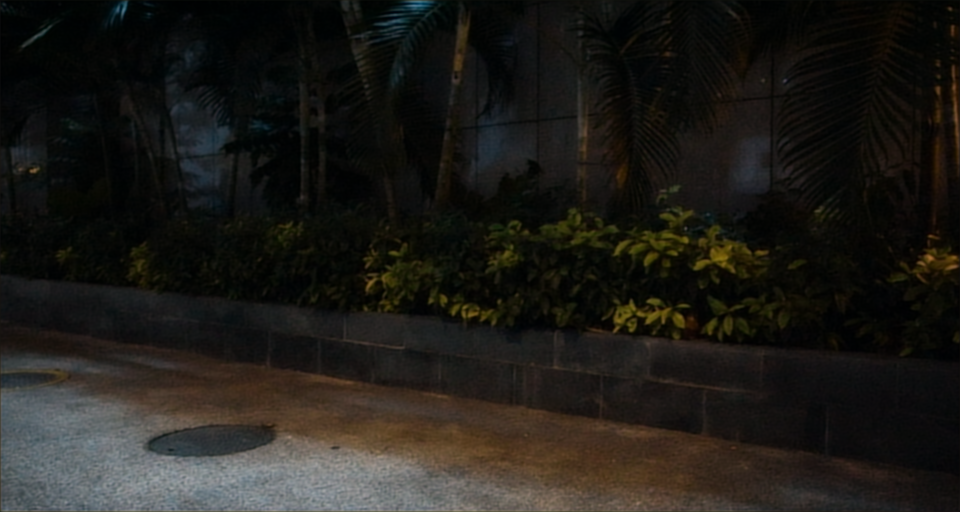}} \hskip3.0em  
\subfloat[Score:26.67, Target] {\includegraphics[width=0.20\linewidth]{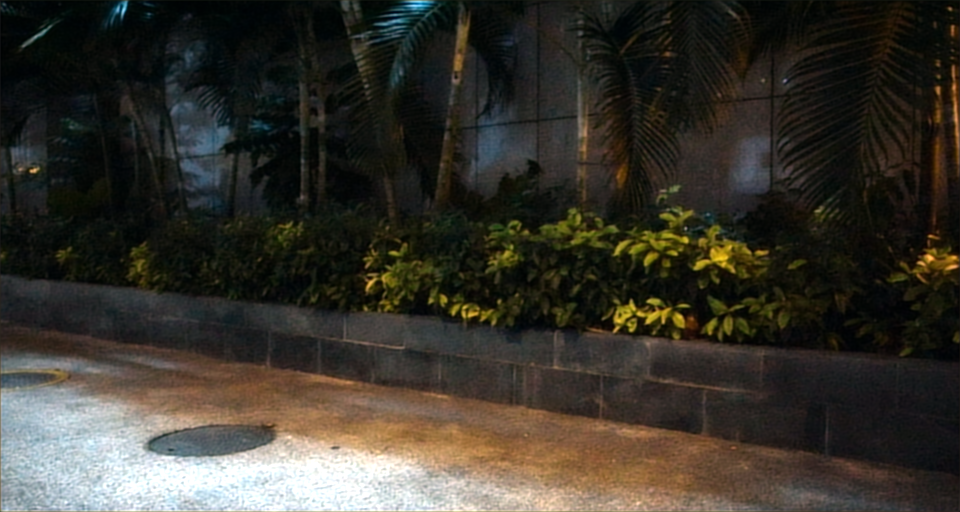}}\hskip3.0em    
\subfloat[Score:33.90] {\includegraphics[width=0.20\linewidth]{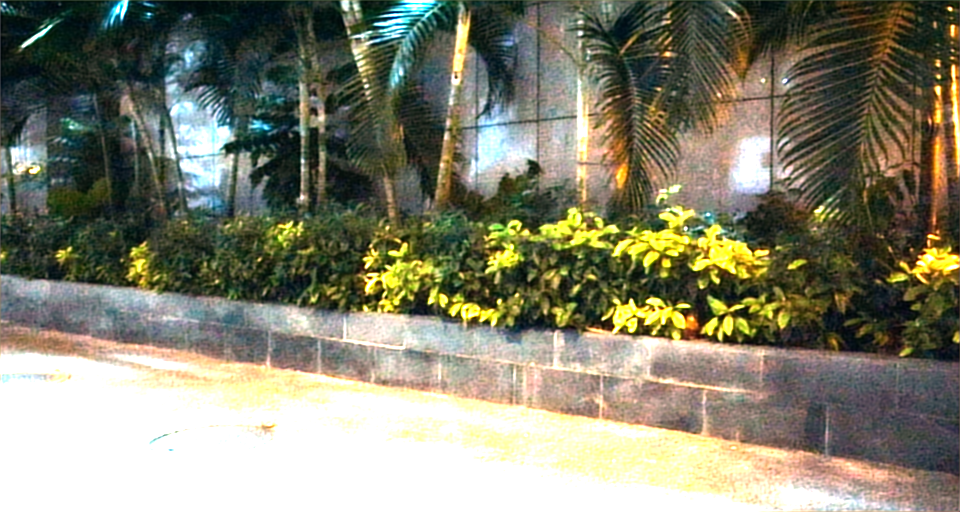}}\hskip3.0em 
\caption{
Illustration of the results based on Eq.~\eqref{gamma correction}.
The proposed method could adaptively select the target based on the BRISQUE score$\downarrow$.}
\label{fig: hsv pseudo reference}
\end{figure*}

Comparatively, achieving illumination styles beyond expert retouching can still be accomplished by integrating the human perception information to precisely control attribute parameters in the HSV color space (brightness, contrast, saturation, and hue) due to the existence of customization of the illumination characteristics according to specific preferences.
As shown in Fig.~\ref{fig: hsv pseudo reference}, the target video frames, denoted as $\boldsymbol{m}_{i}^{k+1}$, are modified through gamma correction and linear scaling as follows, 
\begin{footnotesize}
\begin{equation} \label{gamma correction}
\boldsymbol{m}_{i}^{k+1}=\beta \times(\alpha \times \widetilde{\boldsymbol{x}}_{i}^{k+1})^\gamma.
\end{equation}
\end{footnotesize}
The parameters $\gamma$, $\alpha$, and $\beta$ control the extent of correction and scaling, with values sampled from a uniform distribution range of $U(1.00, 1.10)$.
The efficient quality assessment model BRISQUE~\cite{mittal2012no}~\footnote{\url{https://github.com/chaofengc/IQA-PyTorch.git}} is employed to simulate a proxy human visual system feedback mechanism to select the target video frames, enabling optimization of enhancement algorithms by considering natural scene statistics to avoid under-exposure or over-exposure outcomes.

\noindent \textbf{Loss Function}: The content self-supervised loss it used to preserve the content of images across multiple scales (as shown in Fig.~\ref{main framework} (c)) while allowing for variations in illumination conditions,
\begin{footnotesize}
\begin{equation} \label{content self-supervised loss}
\begin{aligned}
& \mathcal{L}_{\text {content-G}}=\sum_{i={t-2}}^{t+2}\sum_{s=1}^{S}\left\|   \left(G_{\boldsymbol{\theta}}^{k+1}\left(\boldsymbol{y}_i^{k+1}\right) + \boldsymbol{y}_i^{k+1}\right) -  \boldsymbol{m}_{i}^{k+1}     \right\|_1.
\end{aligned}
\end{equation}
\end{footnotesize}

\subsubsection{Temporal Consistency Optimization.} 
We introduce $T_{\boldsymbol{\theta}} \left(\cdot\right)$ to obtain the information from adjacent frames.
The overall unit of temporal learning is defined as,
\begin{footnotesize}
\begin{equation} \label{Inter Net Optimization}
{F}_{\text {Inter}}\left(\widetilde{\boldsymbol{x}}^{k}\right):\left\{\begin{array}{l}
\boldsymbol{r}_{\text{inter}}^k=T_{\boldsymbol{\theta}}^k\left(\widetilde{\boldsymbol{x}}^{k}\right), \boldsymbol{x}^0=\widetilde{\boldsymbol{x}}^{0} \\
\boldsymbol{x}^{k+1}=\widetilde{\boldsymbol{x}}^{k}+\boldsymbol{r}_{\text{inter}}^k,
\end{array}\right.
\end{equation}
\end{footnotesize}
where $\boldsymbol{r}_{\text{inter}}^k$ and $\boldsymbol{x}^{k+1}$ denote the residual and enhanced term. 
In this following description, we only focus on detailing the procedure for one stage, either stage $\textit{k}$ or $\textit{k+1}$ adopting the same procedure.

\noindent \textbf{Stage $\textit{k}$}: 
We broadcast adjacent frames spatially adaptively to correct and complement the information within the current frame for temporal smoothness learning.
The current frame information can be obtained by extracting the guided side information (\textit{i.e.,} optical flow) and warping it to align with the current frame.
To improve the flow estimation accuracy, in the training phase, we finetune the pre-trained optical flow RAFT~\cite{teed2020raft} ($R_{\boldsymbol{\theta}}$ for simplicity) with the video frames.
The simplified optimization function can be defined as,
\begin{footnotesize}
\begin{equation} \label{smoothness fidelity k}
\widetilde{\boldsymbol{x}}^{k+1}=\underset{\boldsymbol{x}}{\operatorname{argmin}} \frac{1}{2}\left\|\boldsymbol{x}-\left(\widetilde{\boldsymbol{x}}^{k}-\rho \nabla F_{\text {Inter}}\left(\widetilde{\boldsymbol{x}}^{k}\right)\right)\right\|_2^2.
\end{equation}
\end{footnotesize}
The model $T_{\boldsymbol{\theta}}$, which utilizes 3D convolution, is trained by leveraging temporal cues, wherein information from adjacent video frames is aggregated to enhance the representation of the current $t$-th frame.
Furthermore, the detail enhancement process is also trained by leveraging temporal cues, ensuring that the estimated frame remains consistent with all of the aligned adjacent video frames during the training process.
The process is defined as follows,
\begin{footnotesize}
\begin{equation} \label{predict current frame residual}
\begin{aligned}
& \boldsymbol{s}_t^{k}=\underbrace{T_{\boldsymbol{\theta}}^k \left( \widetilde{\boldsymbol{x}}_{(t-2) \rightarrow t}^k, \widetilde{\boldsymbol{x}}_{(t-1) \rightarrow t}^k, 
\widetilde{\boldsymbol{x}}_{(t-1) \rightarrow t}^k,
\widetilde{\boldsymbol{x}}_{(t+1) \rightarrow t}^k, \widetilde{\boldsymbol{x}}_{(t+2) \rightarrow t}^k \right)}_{\text {estimate structure signal}} \\
& \boldsymbol{x}_t^{k+1}= \underbrace{T_{\boldsymbol{\theta}}^k \left( \widetilde{\boldsymbol{x}}_{t \rightarrow (t-2)}^k, \widetilde{\boldsymbol{x}}_{t \rightarrow (t-1)}^k, 
\boldsymbol{s}_t^{k},
\widetilde{\boldsymbol{x}}_t^k, 
\widetilde{\boldsymbol{x}}_{t \rightarrow (t+1)}^k \right)}_{\text {estimate compensation detail}} + \boldsymbol{s}_t^{k}.
\end{aligned}
\end{equation}
\end{footnotesize}

\begin{figure*}[t]
\centering    
\subfloat[$\omega$=0.1] {\includegraphics[width=0.20\linewidth]{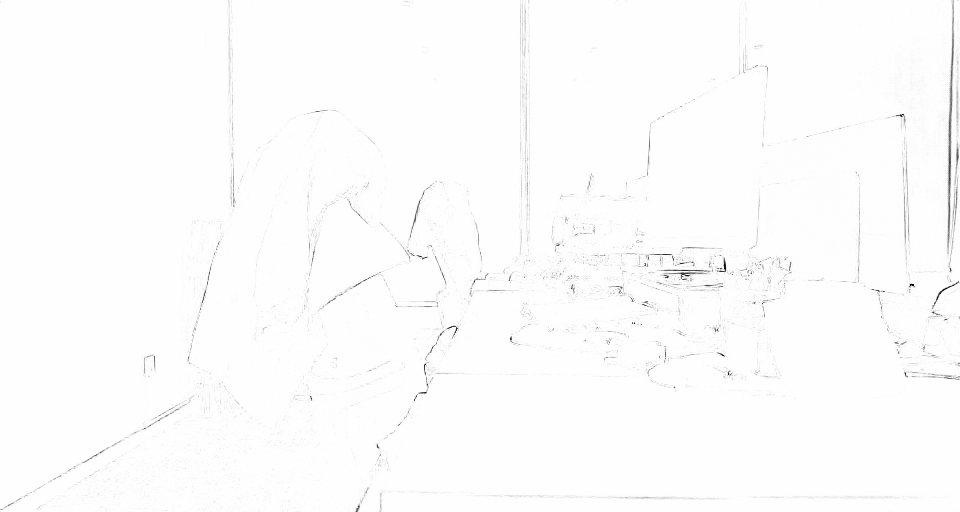}} \hskip3.0em  
\subfloat[$\omega$=0.01] {\includegraphics[width=0.20\linewidth]{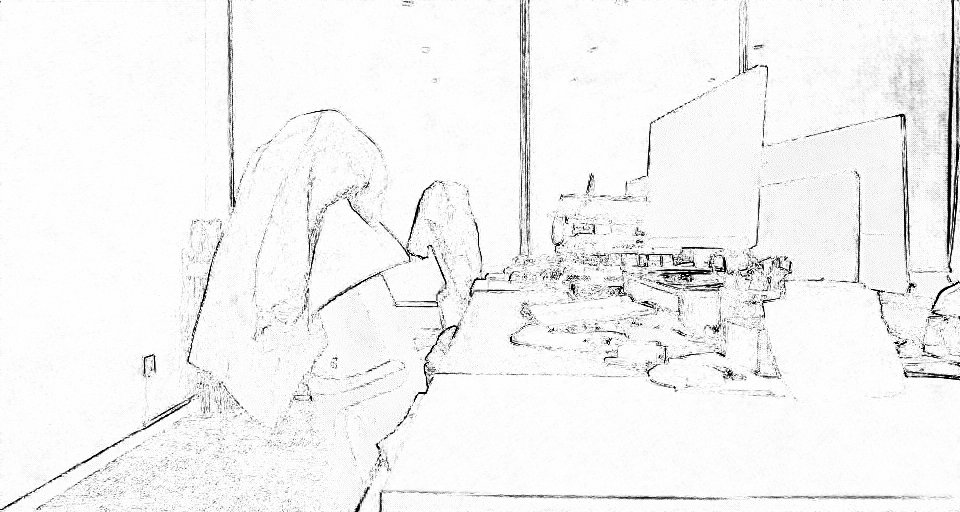}}\hskip3.0em    
\subfloat[$\omega$=0.001] {\includegraphics[width=0.20\linewidth]{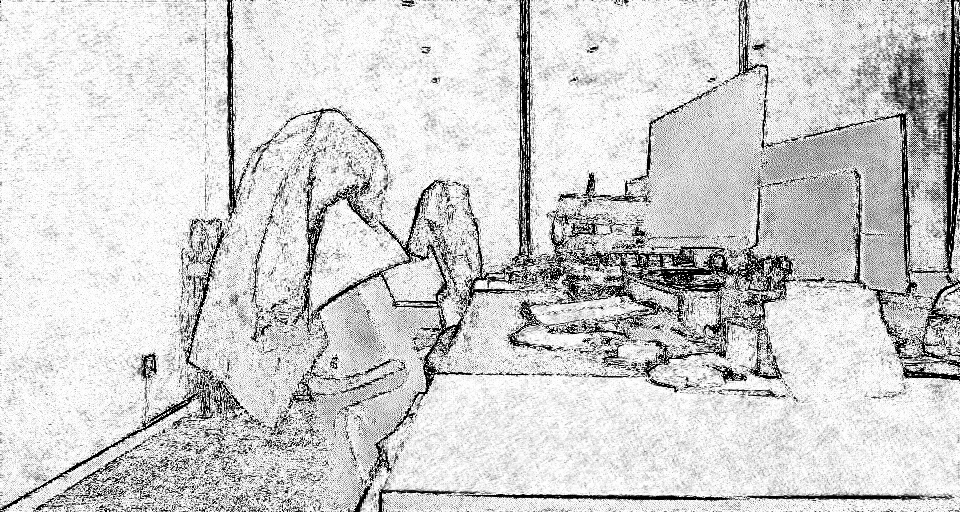}}\hskip3.0em 
\caption{Illusation of the estimated mask results computed using Eq.~\eqref{estimated mask}}
\label{fig: estimated masks}
\end{figure*}

\noindent \textbf{Loss Function}:
The loss function consists of two components: one aims at optimizing the optical flow,
\begin{footnotesize}
\begin{equation} \label{Raft}
\begin{aligned}
& \mathcal{L}_{\text {flow-R}}=\sum_{i={t-2...t+2}/t}\left\|\widetilde{\boldsymbol{x}}_{i \rightarrow t}^k-\widetilde{\boldsymbol{x}}^k_t\right\|_1,
\end{aligned}
\end{equation}
\end{footnotesize}
and the other focuses on optimizing $T_{\boldsymbol{\theta}}(\cdot)$,
\begin{footnotesize}
\begin{equation} \label{T3D}
\begin{aligned}
& \mathcal{L}_{\text {content-T}}=\left\|\boldsymbol{s}^k_t-\widetilde{\boldsymbol{x}}_{t}^k\right\|_1 + \sum_{i=t-2...t+2}\left\|M_i\left(\boldsymbol{x}_{t\rightarrow i}^{k+1}-\widetilde{\boldsymbol{x}}_i^{k}\right)\right\|_1.
\end{aligned}
\end{equation}
\end{footnotesize}
where $M_i$ is the estimated mask~\cite{lai2018learning}.

\subsection{Composite Effect of Different Constraint on Noise Alleviation} 
The proposed method tackles the problem of mixed noise from the following three aspects,

\begin{itemize}
\item \textbf{Spatial Domain Adversarial Learning}. It uses unpaired high-quality noise-free images to make the model reduce the noise present in the generated frames via learning spatial statistics prior.

\item \textbf{Temporal Consistency Learning}. Adjacent frames are aligned and merged to alleviate noise further. 
By aligning the frames using optical flow estimation and warping techniques, the information from neighboring frames is combined and integrated, allowing for noise reduction and enhanced frame quality improvement.

\item \textbf{Mask Mechanism to Balance Noise Suppression and Texture Preservation}. 
Noise removal in low-light images often comes at the expense of texture details.
The parameter $\omega$ controls the shape of the exponential function, as depicted in Fig.~\ref{fig: estimated masks}, which represents the tradeoff between texture and noise.
We can determine the enhanced region $M_t$ of the current frame by utilizing a soft mask that indicates whether pixels are affected by noise or represent the underlying structural signal.
$M_t$ is calculated as follows,
\begin{footnotesize}
\begin{equation} \label{estimated mask}
M_t=\exp \left\{-\frac{\left({\mathrm{ReLU}}\left(\widetilde{\boldsymbol{x}}_t^{k} -\boldsymbol{s}_t^k  \right)\right)^2}{\omega}\right\}.
\end{equation}
\end{footnotesize}

\end{itemize}

\begin{table*}[t]
\centering
\caption{Quantitative results on the SDSD \cite{wang2021seeing} dataset. The best scores are \textbf{highlighted} except for the supervised methods. * and ** denote supervised video-based methods and self-supervised video-based methods, respectively.}
\scriptsize
\begin{tabularx}{\textwidth}{l|*{4}{>{\centering\arraybackslash}X}|*{4}{>{\centering\arraybackslash}X}}\toprule
\multirow{2}{*}{\textbf{Method}}  & \multicolumn{4}{c|}{Outdoor}     & \multicolumn{4}{c}{Indoor}     \\
&PSNR$\uparrow$   & SSIM$\uparrow$    & ${warp}$$\downarrow$   & MABD $\downarrow$ & PSNR$\uparrow$    & SSIM $\uparrow$   & ${warp}$$\downarrow$  & MABD $\downarrow$      \\ \hline\hline
BIMEF~\cite{ying2017bio}     & 18.51    & 0.5572      & 3.15    & 1.54     & 17.91    & 0.6468    & 3.15   & 1.89    \\
Dong~\cite{dong2011fast}     & 13.86    & 0.3575      & 10.16   & 4.93     & 20.55    & 0.4965    & 8.77   & 5.27    \\
LIME~\cite{guo2016lime}      & 9.75     & 0.2783     & 16.83   & 8.35     & 14.92    & 0.4252     & 16.45  & 10.55   \\
MF~\cite{fu2016fusion}       & 14.99    & 0.4113     & 8.06    & 3.89     & 20.65    & 0.5620     & 6.63   & 3.98    \\
MR~\cite{jobson1997multiscale}   & 8.10     & 0.2805    & 20.12   & 8.93     & 10.25    & 0.4244    & 22.05  & 11.26    \\
NPE~\cite{wang2013naturalness}     & 12.08    & 0.3399    & 14.90   & 6.69     & 16.72    & 0.4664   & 12.65  & 6.34    \\
SRIE~\cite{fu2016weighted}         & 21.89    & 0.6288     & 2.74    & 1.42     & 15.78    & 0.6294   & 2.75   & 1.81    \\\hline
EnlightenGAN~\cite{jiang2021enlightengan}  & 18.63    & 0.5399    & 4.49    & 2.52     & 19.59    & 0.5874  & 3.37   & 2.48    \\
RUAS~\cite{liu2021retinex}       & 11.83    & 0.4000    & 3.42    & 2.14     & 20.54    & 0.6071   & 2.31   & 3.13    \\
SCI~\cite{ma2022toward}         & 17.35    & 0.4651    & 3.53    & 1.87     & 13.69    & 0.6189     & 0.77    & 0.84    \\
ZeroDCE~\cite{guo2020zero}      & 6.54     & 0.2081    & 20.80   & 9.05     & 13.27    & 0.4631   & 16.41  & 8.68    \\
CLIP-LIT~\cite{liang2023iterative}   & 20.88 & 0.5872  & 3.36 & 1.85  & 19.08  & 0.4582  & 11.75  &  6.72    \\
SGZSL~\cite{zheng2022semantic}   & 6.09     & 0.1899     & 19.47   & 8.48     & 14.38    & 0.4793    & 12.34  & 6.89  \\\hline
MBLLVEN*~\cite{lv2018mbllen}    & 16.38    & 0.5573     & 4.76    & 2.03     & 23.78    & 0.7845     & 0.79   & 1.90    \\
DRVNet*~\cite{chen2019seeing}   & 17.39    & 0.6656     & 1.41    & 0.57     & 26.11    & 0.8518    & 0.45   & 1.20    \\
StableLLVE*~\cite{zhang2021learning} & 20.10    & 0.7510   & 4.84    & 1.73     & 24.76    & 0.8369  & 1.63   & 1.73    \\
SDSDNet*~\cite{wang2021seeing}     & 24.30    & 0.7445   & 0.95    & 0.47     & 27.03    & 0.7788   & 1.74   & 2.03    \\\hline
PSENet**~\cite{nguyen2023psenet}     & 11.75 & 0.3541  & 10.00 & 4.81  & 17.79  & 0.5459  & 7.84  &  5.10    \\
SALVE**~\cite{azizi2022salve}      & 18.72    & 0.5888     & 1.09   & 0.48     & 17.09   & 0.7215     & 0.94   & \textbf{0.64}    \\\hline
Ours      & \textbf{23.94} & \textbf{0.7446} & \textbf{0.24} & \textbf{0.21} & \textbf{22.41} & \textbf{0.7368}  & \textbf{0.41}  & {1.05} \\ \bottomrule

\end{tabularx}
\label{tab: video enhancement result SDSD}
\end{table*}

\begin{figure*}[t]
\centering
\includegraphics[width=0.8\linewidth]{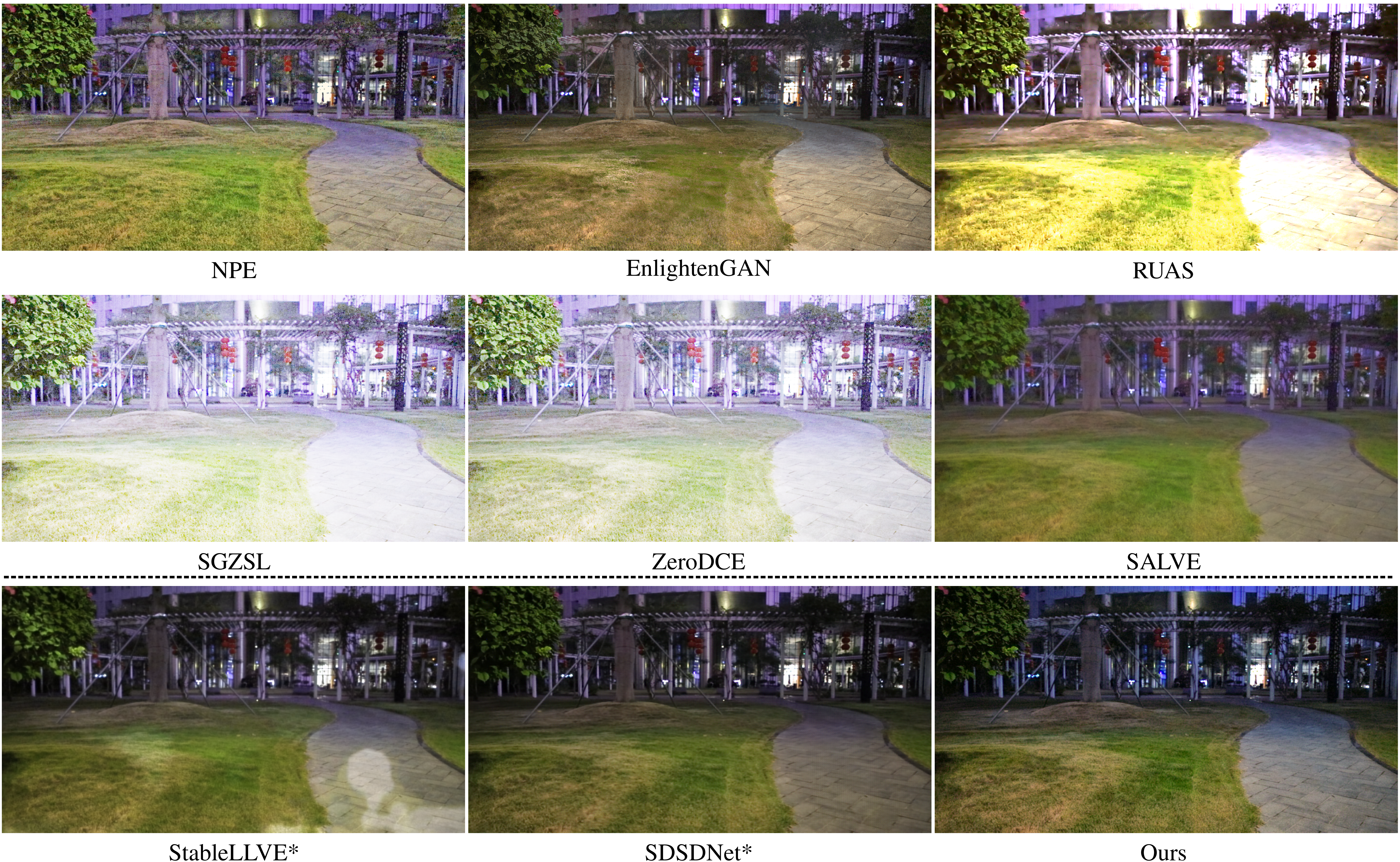}
\caption{Qualitative comparison results on SDSD outdoor dataset~\cite{wang2021seeing}.
The supervised methods (*) are provided below the dashed line for reference.}
\label{fig: SDSD outdoor 76}
\end{figure*}

\section{Experiments}
We present the implementation details, ablation analysis, and comparison with existing methods. 
The supplementary material provides additional details, including benchmark descriptions, visual comparisons, and network architectures.

\subsection{Implementation Details}
\noindent \textbf{Dataset.}
The proposed method is trained on the SDSD dataset~\cite{wang2021seeing}, which includes realistic outdoor and indoor scenes.
The outdoor scenes include 67 training and 13 testing videos, while the indoor scenes include 58 training and 12 testing videos.

\noindent \textbf{Evaluation Metric.}
We assess the performance of the proposed method using several full-reference quality measures, such as PSNR and SSIM \cite{wang2004image}, which offer valuable insights into the effectiveness, with higher values indicating better reconstruction quality.
In addition, we report ${warp}$~\cite{lai2018learning}, and MABD~\cite{jiang2019learning} value to assess the temporal smoothness, where lower values indicate better quality in terms of temporal smoothness.
%a

\begin{figure*}[t]
\centering
\includegraphics[width=0.8\linewidth]{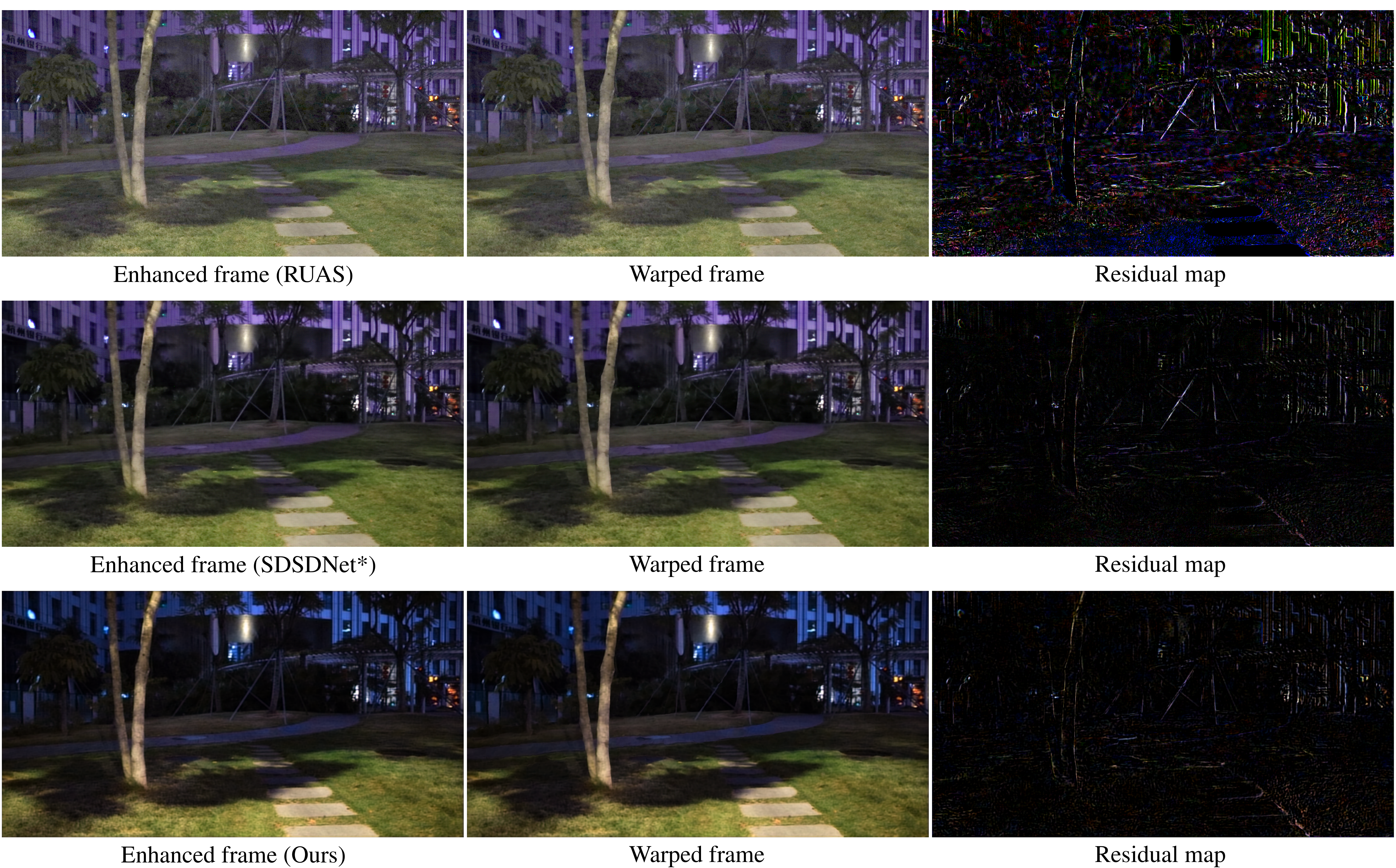}
\caption{Qualitative comparison results on SDSD outdoor dataset~\cite{wang2021seeing}.
First column: enhanced results. Second column: warped results. Third column: per-pixel warping error.
}
\label{fig: SDSD outdoor 75 temporal error}
\end{figure*}

\subsection{Comparison to the State-of-the-arts}
\noindent\textbf{Quantitative Evaluation.}  
The comprehensive evaluation of the proposed method demonstrates promising results over the second-best-performing reference-free method, indicated by significant improvements across quality measures, including PSNR, SSIM, ${warp}$, and MABD, as shown in Table~\ref{tab: video enhancement result SDSD}.
Specifically, it achieves a gain of 2.05 dB and 0.1158 SSIM value on the outdoor scenes and shows a gain of 1.76 dB and 0.0153 SSIM value on the indoor scenes.
On the outdoor dataset, the proposed method achieves results comparable to supervised methods like MBLLVEN, DRVNet, StableLLVE, and SDSDNet, specifically achieving an SSIM value of 0.7446.
In addition, the proposed method exhibits promising performance in terms of temporal quality measures compared to other methods. 
Specifically, it exhibits superior performance compared to the second-based methods, with a reduction of 0.85 in ${warp}$ error and 0.27 in MABD on outdoor scenes, as well as a reduction of 0.53 in ${warp}$ error on indoor scenes.
These findings highlight the effectiveness of the proposed method in accurately handling temporal variations and producing high-quality smoother results in outdoor and indoor scenes.

\noindent\textbf{Qualitative Evaluation.}
By qualitatively comparing the results in Fig.~\ref{fig: SDSD outdoor 76} and the temporal warping results in Fig.~\ref{fig: SDSD outdoor 75 temporal error}, which include both single-image and video-based enhancement methods, several observations can be made.  
In Fig.~\ref{fig: SDSD outdoor 76}, it is evident that the proposed method excels in preserving essential structural details, maintaining high contrast, and reducing noise artifacts, which enhance the overall video frame quality.
Notably, the proposed method outperforms the supervised method StableLLVE and is comparable to the SOTA-supervised method SDSDNet.
In comparison, the other enhancement methods still struggle with severe artifacts such as persistent noise, color casting, and abnormal results.
Specifically, it is observed that methods like EnlightenGAN and SALVE struggle with color-casting issues. 
RUAS, SGZSL, and ZeroDCE tend to introduce over-exposure in their results. 
The qualitative evaluation presented in Fig.~\ref{fig: SDSD outdoor 75 temporal error} serves the purpose of analyzing the temporal smoothness between adjacent frames. 
By examining the residual map, it can be concluded that the proposed method surpasses other enhancement methods and is comparable to the supervised method SDSDNet, demonstrating superior performance in maintaining temporal consistency.

\subsection{Ablation Study}
\begin{itemize}

\item Ours-v1: Trained with only subnetwork (a) for illumination learning in the first stage.
\item Ours-v2: Trained with subnetwork (a) for illumination learning and subnetwork (b) for temporal smoothness learning in the first stage.
\item Ours-v3: In stage 2, an additional subnetwork (c) is introduced for illumination learning.
\item Default: Incorporates subnetworks (a), (b), (c), and (d) and involves human perception feedback in stage 2.
This version is the final result.

\item  Default w.o H: Incorporates subnetworks (a), (b), (c), and (d), but does not include the Human Perception Feedback Mechanism.

\end{itemize}

\begin{table*}[t]
\centering
\scriptsize
\caption{Ablation study of each component.}
\begin{tabular}{l|cccc|cccc} 
\toprule
\multirow{2}{*}{\textbf{Network}}  & \multicolumn{4}{c|}{Outdoor}     & \multicolumn{4}{c}{Indoor}            \\
& PSNR$\uparrow$   & SSIM$\uparrow$   & ${warp}$$\downarrow$   & MABD $\downarrow$ & PSNR$\uparrow$    & SSIM $\uparrow$   & ${warp}$$\downarrow$  & MABD $\downarrow$      \\\hline\hline
Ours-v1 (stage=1)    & 21.29 & 0.6555    & 0.54   & 0.36   & 21.73 & 0.6968    &  1.07     & 1.25   \\
Ours-v2 (stage=1)    & 21.43 & 0.6749   & 0.20    & 0.16   & 21.90 & 0.7417    & 0.40      & 0.86 \\
Ours-v3 (stage=2)    & 23.92 & 0.7455   & 0.35   & 0.29   & 22.43 & 0.7448   & 0.51     & 1.15 \\
Default w.o H & 21.87  & 0.6824     &  0.14   & 0.13  & 22.00 & 0.7356   & 0.33   & 0.81 \\\hline
Default (stage=2)       & 23.94 & 0.7446  & 0.24 & 0.21  & 22.41  & 0.7368  & 0.41  &  1.05 \\\bottomrule

\end{tabular}%
\label{tab: network component ablation}%
\end{table*}%

\begin{table*}[t]
\centering

\caption{Quantitative results on parameters and inference time.
}
\label{tab: Model complexity}
\scriptsize
\begin{tabularx}{\textwidth}{>{\centering\arraybackslash}Xcc|>{\centering\arraybackslash}Xcc|>{\centering\arraybackslash}Xcc}
\toprule
Model  &  Parameters (M)  & Time (ms) & Model  &  Parameters (M)  &  Time (ms) & Model  &  Parameters (M)  &  Time (ms)  \\
\midrule
$G \left(\cdot\right)$     & 4.16 & 3.99  & $D \left(\cdot\right)$      & 4.63 & - & $T \left(\cdot\right)$  & 0.30 & 1.68\\
\bottomrule
\end{tabularx}
\end{table*}

\noindent \textbf{Network Components.}
Table~\ref{tab: network component ablation} provides insights into the contributions of each network in the proposed method, allowing for analysis of the underlying mechanisms and offering good interpretability.
In stage 1, it can be concluded that subnetwork (a) contributes to the recovery of illumination to some extent, while subnetwork (b) enhances the video by making it smoother. 
Similarly, in stage 2, subnetwork (c) and subnetwork (d) have similar roles to subnetwork (a) and subnetwork (b), respectively.

\noindent \textbf{Human Perception Feedback Mechanism.}
The results of Default w.o H in Table~\ref{tab: network component ablation} demonstrate that incorporating the generated reference image based on human perception feedback (Eq.~\eqref{gamma correction}) significantly improves the overall visual quality.

\subsection{Limitations and Model Complexity}
Further improvements can be achieved by investigating more advanced techniques like diffusion-based methods. 
In addition, if computational resources are allowed, it would be beneficial to use a more powerful quality assessment model as human-designed feedback instead of traditional BRISQUE~\cite{mittal2012no}.
Table~\ref{tab: Model complexity} presents the model complexity in terms of model size and inference time for a spatial input frame size of $960 \times 512$, with an acceptable inference time of 3.99 ms for $G(\cdot)$ and 1.68 ms for $T(\cdot)$.

\section{Conclusion}
In this paper, we propose a reference-free strategy for addressing low-light video enhancement in various real-world scenarios.
The proposed UDU-Net stands out from previous methods in low-light video enhancement by leveraging a progressive spatial-temporal optimization perspective. 
In terms of spatial optimization, the proposed method incorporates expert-retouched illuminations and introduces controllable human-perception feedback to enhance video quality, refining the illumination from multi-granularity representation.
In the temporal optimization aspect, the proposed method makes use of temporal correlation cues to map low-light video frames to normal-light frames, reconstructing main structures and estimating structural details.
Through comprehensive exploration of the properties of the proposed method and conducting extensive experiments, we have successfully demonstrated the effectiveness and superiority of the method in enhancing low-light videos under indoor and outdoor scenes.

\clearpage

% \section*{Acknowledgements} 
% %
% This work was supported in part by the Hong Kong Innovation and Technology Commission (InnoHK Project CIMDA), in part by the General Research Fund of the Research Grant Council of Hong Kong under Grants 11203220 and 11200323, in part by ITF Project GHP/044/21SZ, in part by (Guangdong Basic and Applied Basic Research Foundation) (2024A1515010454), in part by the Basic and Frontier Research Project of PCL, in part by the Major Key Project of PCL, in part by the Natural Science Foundation of China under Grant 62201387, in part by the Shanghai Pujiang Program under Grant 22PJ1413300, and in part by the National Natural Science Foundation of China under Grant 62201526.

% \clearpage\mbox{}Page \thepage\ of the manuscript.
% \clearpage\mbox{}Page \thepage\ of the manuscript.
% \clearpage\mbox{}Page \thepage\ of the manuscript.
% \clearpage\mbox{}Page \thepage\ of the manuscript.
% \clearpage\mbox{}Page \thepage\ of the manuscript. This is the last page.
% Now we have reached the maximum length of an ECCV \ECCVyear{} submission (excluding references).
% References should start immediately after the main text, but can continue past p.\ 14 if needed.
% \clearpage  % TODO REVIEW/FINAL: This \clearpage needs to be removed from both review and camera-ready versions.

\bibliographystyle{splncs04}
\bibliography{reference}
\end{document}